\documentclass[10pt,twocolumn,letterpaper]{article}

\usepackage[pagenumbers]{cvpr}

\usepackage{graphicx}
\usepackage{amsmath}
\usepackage{amssymb}
\usepackage{booktabs}
\usepackage[dvipsnames]{xcolor}
\usepackage{pifont}
\usepackage{placeins}
\usepackage{makecell}

\DeclareMathOperator*{\argmax}{argmax}
\usepackage[accsupp]{axessibility}

\definecolor{cvprblue}{rgb}{0.21,0.49,0.74}
\usepackage[pagebackref,breaklinks,colorlinks,allcolors=cvprblue]{hyperref}
\usepackage[capitalize]{cleveref}
\crefname{section}{Sec.}{Secs.}
\Crefname{section}{Section}{Sections}
\Crefname{table}{Table}{Tables}
\crefname{table}{Tab.}{Tabs.}
\def\confName{CVPR}
\def\paperID{12}
\def\confYear{2025}

\begin{document}

\title{\textcolor[HTML]{349ccc}{ITA}CLIP: Boosting Training-Free Semantic Segmentation with \textcolor[HTML]{349ccc}{I}mage, \textcolor[HTML]{349ccc}{T}ext, and \textcolor[HTML]{349ccc}{A}rchitectural Enhancements}

\author{M. Arda Aydın \\
Bilkent University \\
{\tt\small arda.aydin@bilkent.edu.tr}
\and
Efe Mert Çırpar\\
RWTH Aachen University\\
{\tt\small efe.cirpar@rwth-aachen.de}
\and
Elvin Abdinli\\
Technical University of Munich\\
{\tt\small elvin.abdinli@tum.de}
\and
Gozde Unal\\
Istanbul Technical University\\
{\tt\small gozde.unal@itu.edu.tr}
\and
Yusuf H. Sahin\\
Istanbul Technical University\\
{\tt\small sahinyu@itu.edu.tr}
}
\maketitle

\begin{abstract}
   Recent advances in foundational Vision Language Models (VLMs) have reshaped the evaluation paradigm in computer vision tasks. These foundational models, especially CLIP, have accelerated research in open-vocabulary computer vision tasks, including Open-Vocabulary Semantic Segmentation (OVSS). Although the initial results are promising, the dense prediction capabilities of VLMs still require further improvement. In this study, we enhance the semantic segmentation performance of CLIP by introducing new modules and modifications: 1) architectural changes in the last layer of ViT and the incorporation of attention maps from the middle layers with the last layer, 2) Image Engineering: applying data augmentations to enrich input image representations, and 3) using Large Language Models (LLMs) to generate definitions and synonyms for each class name to leverage CLIP's open-vocabulary capabilities. Our training-free method, ITACLIP, outperforms current state-of-the-art approaches on five popular segmentation benchmarks. Our code is available at \url{https://github.com/m-arda-aydn/ITACLIP}.
   
\end{abstract}

\section{Introduction}
\label{sec:intro}
The emergence of large foundational Vision Language Models (VLMs) \cite{alayrac2022flamingo,ilharco_gabriel_2021_5143773,jia2021scaling,radford2021learning} has driven a paradigm shift in deep learning. Before the introduction of these foundational models, computer vision models were effective only for a limited number of classes. However, this evaluation setting does not accurately reflect real-world conditions, as the world is not limited to a small number of classes, and models trained on these finite classes exhibit weak transferability to real-world scenarios. These reasons highlight the need for open-vocabulary models rather than relying on predefined classes.

\begin{figure}
    \centering
    \includegraphics[width=1\linewidth]{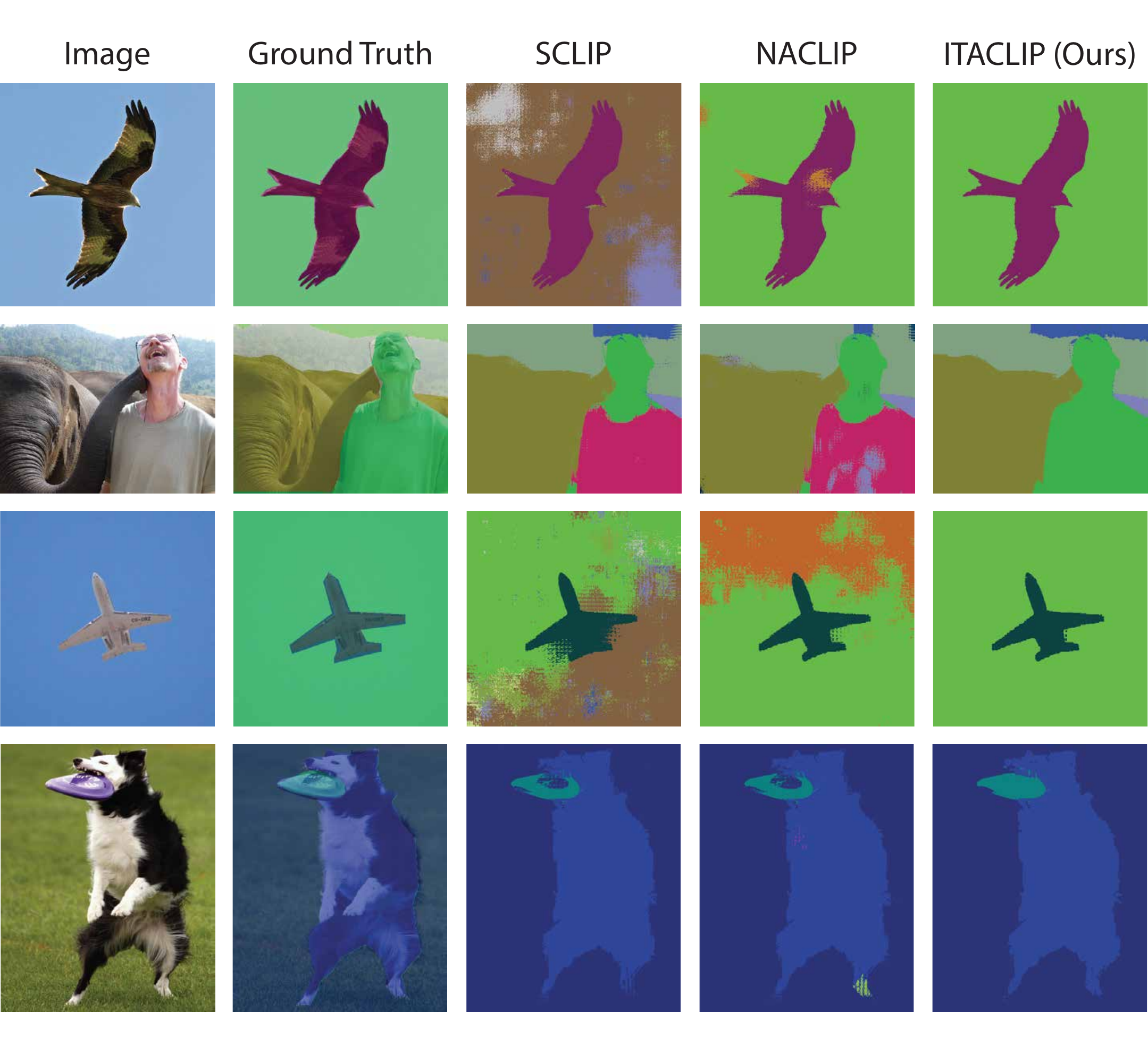}
    \caption{\textbf{Qualitative comparison of training-free semantic segmentation methods.} We compare ITACLIP with SCLIP \cite{wang2023sclip} and NACLIP \cite{hajimiri2024pay} using images from the COCO-Stuff \cite{caesar2018coco} dataset. Additional visualizations are included in the Appendix.}
    \label{fig:fig_1}
\end{figure}

Rapid advancements in VLMs have fueled research on open-vocabulary computer vision tasks. In particular, the CLIP \cite{radford2021learning} model has become foundational for various computer vision tasks such as multi-label image classification \cite{abdelfattah2023cdul,lin2024tagclip}, object detection \cite{guopen,shi2023edadet,du2022learning}, semantic segmentation \cite{lin2023clip,shao2024explore,barsellotti2024fossil,zhou2023zegclip,li2023clip,li2024cascadeclip,ranasinghe2023perceptual,shin2022reco,lan2024clearclip,sun2023going}, and image generation \cite{tao2023galip,wang2022clip}. The generalization capability of CLIP has enabled models to achieve remarkable results in open-vocabulary settings with relatively minimal modifications instead of requiring training from scratch.

Despite the great success of VLMs in zero-shot image classification \cite{radford2021learning,jia2021scaling}, translating this achievement into dense prediction tasks has become challenging for researchers. Although several works \cite{zhou2023zegclip,ding2022decoupling,xu2022simple} have demonstrated strong results in zero-shot semantic segmentation by employing new components, these models still need expensive \textit{pixel-level} annotations of seen classes. 

Segment Anything Model (SAM) \cite{kirillov2023segment}, a large foundational segmentation model, is trained on a massive dataset and delivers impressive results in semantic segmentation. Yet, the segmentation capability of SAM strongly depends on the contents of its training dataset, and further studies \cite{mazurowski2023segment,he2023computer,deng2023segment} indicate that its performance degrades when applied to images dissimilar to those in the training data. Furthermore, collecting annotated data can be quite expensive in certain domains, such as biomedical image analysis, where labeling requires specialized expertise. This limitation significantly impacts the scalability of models. In response to these challenges, researchers have turned their focus to training-free semantic segmentation models as a promising solution. Training-free semantic segmentation models typically use a VLM like CLIP and aim to transform VLM's image-level knowledge into pixel-level predictions for dense prediction tasks \cite{li2023clip,bousselham2024grounding,wang2023sclip,hajimiri2024pay,wysoczanska2024clip,sun2024clip,zhou2022extract,barsellotti2024fossil}.

In this study, we present \mbox{\textbf{ITACLIP}} (\textbf{I}mage-\textbf{T}ext-\textbf{A}rchitectural Enhanced \textbf{CLIP}), our training-free method, which utilizes architectural changes and enriches the input features to improve the segmentation performance. Our approach outperforms the current state-of-the-art methods on COCO-Stuff \cite{caesar2018coco}, COCO-Object \cite{lin2014microsoft}, Pascal Context \cite{mottaghi2014role}, Pascal VOC \cite{everingham2010pascal}, and Cityscapes \cite{cordts2016cityscapes}. As illustrated in \cref{fig:fig_1}, ITACLIP produces more accurate segmentation maps than SCLIP \cite{wang2023sclip} and NACLIP \cite{hajimiri2024pay}.  

Our contributions can be summarized as follows: 
 
\begin{itemize}
    \item We employ a combination of \textit{self-self attentions} instead of the original attention mechanism of ViT \cite{dosovitskiy2020image}. Moreover, we \textit{remove} the Feed-Forward Network (FFN) in the last layer of ViT and combine the \textit{attention maps} from the middle layers with the attention map of the final layer.
    \item We leverage Large Language Models (LLMs) to develop a systematic strategy for \textit{generating auxiliary texts} for any class name in an open-vocabulary setting. We integrate text features from the original class names with those from definitions or synonyms.
    \item We propose the \textit{Image Engineering} module, a novel component designed to refine features extracted from the image encoder. Rather than feeding only the original image to CLIP's visual encoder, we also utilize augmented images to enrich and diversify the overall image representation.
    \item We present a comprehensive analysis demonstrating that ITACLIP achieves state-of-the-art results on various segmentation benchmarks.

\end{itemize}
\section{Related Work}
\label{sec:related work}

\subsection{Foundational Vision-Language Models}
Inspired by the success of large-scale foundational models in natural language processing \cite{devlin2018bert, raffel2020exploring, radford2018improving,peters-etal-2018-deep}, researchers have turned their attention to training foundational models in computer vision. As part of this effort, VLMs aim to integrate textual and visual information to construct a unified understanding. Contrastive learning-based VLMs have demonstrated impressive capabilities in visual-language understanding \cite{radford2021learning,jia2021scaling,ilharco_gabriel_2021_5143773,alayrac2020self,cherti2023reproducible,alayrac2022flamingo,li2021align,xu2023demystifying}.

Contrastive Language-Image Pre-training (CLIP) \cite{radford2021learning} includes both an image encoder and a text encoder, jointly trained on a private WIT-400M dataset containing image-text pairs. CLIP has exhibited strong transferability in zero-shot visual recognition tasks, fundamentally changing the paradigm for zero-shot learning in computer vision. Like CLIP, ALIGN \cite{jia2021scaling} employs a dual-encoder architecture; however, it is trained on a much larger private dataset to enhance scalability and simplify dataset curation. OpenCLIP \cite{ilharco_gabriel_2021_5143773} is an open-source implementation of CLIP and is trained on the publicly available LAION \cite{schuhmann2022laion} dataset. Due to the strong transferability of CLIP on downstream tasks, many training-free semantic segmentation models \cite{lan2024clearclip,wang2023sclip} employ OpenAI's pre-trained CLIP model. Hence, our method also follows this implementation.

\subsection{Open-Vocabulary Semantic Segmentation}

Semantic segmentation can be defined as the pixel-wise classification of an image. Conventionally, semantic segmentation models are trained on a limited set of known classes, and the evaluation process is conducted for these specific classes. In contrast, open-vocabulary semantic segmentation models use VLMs to assign \textit{arbitrary} class labels to each pixel in a given image. The target class set includes not only predefined classes but also arbitrary class names. Open-vocabulary semantic segmentation models can be categorized into three groups: 1) \textbf{Training-Free} Methods, 2) \textbf{Weakly Supervised} Methods, and 3) \textbf{Fully Supervised} Methods.

\begin{figure*}
    \centering
    \includegraphics[width=0.95\linewidth]{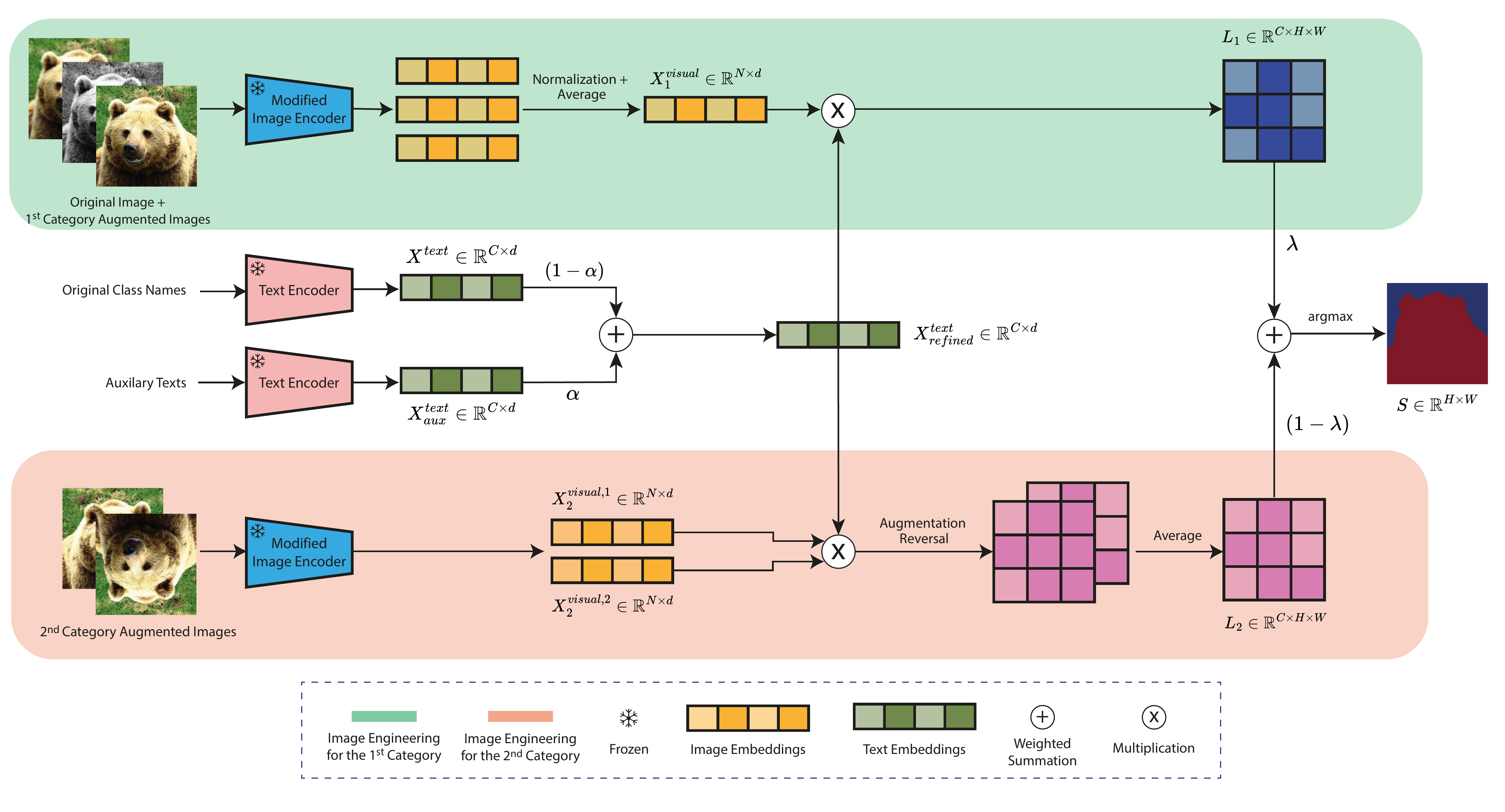}
    \caption{\textbf{Overview of ITACLIP.} Our method integrates image, text, and architectural enhancements to produce a more accurate segmentation map. We apply various data augmentation techniques, then process both the original and augmented images through a modified image encoder to obtain image embeddings. We also utilize an LLM to generate auxiliary texts (\eg, definitions or synonyms) for each original class name. The $\lambda$ and $\alpha$ symbols denote the image engineering and auxiliary text coefficients used in weighted summations, respectively.}
    \label{fig:overview}
\end{figure*}

\textbf{Training-Free Methods.} Training-free semantic segmentation models make predictions without using pixel-level or image-level annotations. MaskCLIP \cite{zhou2022extract} discards the query and key embeddings, using only the value embedding from the final self-attention block while modifying the last linear layer of CLIP's visual encoder. Inspired by MaskCLIP, research on training-free semantic segmentation models has rapidly accelerated. SCLIP \cite{wang2023sclip} introduces a new attention formula based on query-query and key-key attention, leading to improved performance compared to MaskCLIP. GEM \cite{bousselham2024grounding} employs generalized self-self attention combinations along with certain regularization steps. NACLIP \cite{hajimiri2024pay} incorporates a Gaussian window into the attention map of each patch to increase spatial attention around the patch and modifies CLIP's visual encoder to boost the model's performance. CLIP-DIY \cite{wysoczanska2024clip} extracts patch-level features from various image crops and aggregates them before upsampling. Alternatively, some methods \cite{lin2024tagclip, sun2024clip} utilize Class Activation Maps (CAMs) \cite{zhou2016learning} to localize regions activated by the target class and then perform segmentation based on these activated areas. TagCLIP \cite{lin2024tagclip} first performs multi-label classification using modules designed for classification and feeds the predicted classes to CLIP-ES \cite{lin2023clip}, a CAM-based segmentation framework, for segmentation. CaR \cite{sun2024clip} operates in two-stage units and recurrently generates masks until the predicted classes remain unchanged. The two-stage unit comprises two components: a mask generator (using CLIP-ES) and a mask classifier (using CLIP).

\textbf{Weakly Supervised Methods.} Weakly supervised semantic segmentation (WSSS) models \cite{luo2023segclip,xu2023learning,xu2022groupvit,cha2023learning,ranasinghe2023perceptual,lin2023clip} rely on weak supervision, such as image-level labels, rather than labor-intensive pixel-level annotations. GroupViT \cite{xu2022groupvit} learns visual representations by grouping semantically related regions within images. TCL \cite{cha2023learning} leverages text-grounded images and introduces a ``text-grounded contrastive loss'' to align captions with corresponding regions. OVSegmentor \cite{xu2023learning} uses Slot Attention \cite{locatello2020object} to group visual features and aligns these groups with related text features.

\textbf{Fully Supervised Methods.} Fully supervised semantic segmentation models  \cite{han2023open,jiao2023learning,liang2023open,ghiasi2022scaling,lilanguage,li2023open,xu2023open} are typically trained on a specified dataset with pixel-level annotations available. Since these models have access to fine-grained dense labels, they generally outperform training-free and weakly supervised models.

Our study focuses on training-free semantic segmentation, aiming to extend CLIP’s image-level capabilities to the pixel-level without relying on additional training or external models trained on \textit{pixel-level} annotations, like SAM. Thus, ITACLIP does not require pixel-level annotations at any stage to produce segmentation maps.
\section{Method}
\label{sec:method}
In this section, we introduce our training-free semantic segmentation model, ITACLIP, as depicted in \cref{fig:overview}. First, we revisit the original ViT-based CLIP image encoder\footnote{Note that we focus exclusively on ViT-based \cite{dosovitskiy2020image} image encoders due to the limited zero-shot capabilities of ResNet-based \cite{he2016deep} encoders in dense prediction tasks, as demonstrated in \cite{zhou2022extract}.}. Next, we begin our investigation by examining a vanilla approach—a straightforward method utilizing patch tokens for segmentation—as our baseline study. Finally, we present the architectural modifications to the CLIP's ViT and introduce the proposed modules designed to expand the input representation.

\subsection{Revisiting Original CLIP}
\label{revisiting clip}
The original ViT-based CLIP \cite{radford2021learning} consists of two distinct transformer-based encoders: an image encoder and a text encoder. The image encoder takes an input image $I \in \mathbb{R}^{3 \times H \times W}$ and divides it into non-overlapping patches of size $P \times P$, where $H$ and $W$ represent the height and width of the image, respectively. Thus, we obtain $N$ different patch tokens, where $N = HW/P^2$ is the total number of patches. Subsequently, a linear projection maps each patch token to a $d$-dimensional space, with $d$ denoting the feature dimension of the model. Additionally, the class token, \texttt{[CLS]}, is concatenated with the patch tokens extracted from image patches to capture the global context of the image. Lastly, positional embeddings are added to the visual tokens to prepare them as input for the first layer of the encoder.

The $l$th layer of the visual encoder receives visual tokens \mbox{$X^{(l-1)} = [x_{\texttt{CLS}}, x_{1}, \dots, x_{N}] \in \mathbb{R}^{(N + 1) \times d}$} from the previous layer. Next, $X^{(l)}$ is calculated as,

\begin{equation}
  X^{*} = X^{(l-1)} + \texttt{{\textbf{SA}}}(LN(X^{(l-1)}))
  \label{eq:eq_1}
\end{equation}

\begin{equation}
  X^{(l)} = X^{*} + \texttt{{\textbf{MLP}}}(LN(X^{*}))
  \label{eq:eq_2}
\end{equation}

\noindent where ${LN}$, \texttt{{\textbf{SA}}}, and \texttt{{\textbf{MLP}}} represent Layer Normalization, the Self-Attention module, and the Feed-Forward Network (FFN), respectively.

Given an input $X \in \mathbb{R}^{(N + 1) \times d}$, the Self-Attention module can be mathematically described as,
\begin{equation}
  \texttt{\textbf{Attn}}(X) = softmax(\frac{XW_{Q}W_{K}^{T}X^{T}} {\sqrt{d}})
  \label{eq:eq_3}
\end{equation}

\begin{equation}
  \texttt{{\textbf{SA}}}(X) = \texttt{\textbf{Attn}}(X)XW_{V}W_{O}
  \label{eq:eq_4}
\end{equation}

\noindent where $\texttt{\textbf{Attn}}(X) \in \mathbb{R}^{(N + 1) \times (N + 1)}$ denotes the attention map for input $X$, and $W_{Q}, W_{K}, W_{V}, W_{O} \in \mathbb{R}^{d \times d}$ represent query, key, value, and output projection matrices learned during pre-training, respectively. For simplicity, we consider only single-head self-attention in our self-attention module description.

\subsection{Vanilla Approach}
\label{vanilla}
Applying a patch-based classification strategy is the most straightforward approach to utilize CLIP for segmentation. We employ CLIP's image encoder to generate visual features \mbox{$X^{visual} = [x^{visual}_{\texttt{CLS}}, X^{visual}_{patch}] \in \mathbb{R}^{(N + 1) \times d}$} of given image $I$, as described in \cref{revisiting clip}. $x^{visual}_{\texttt{CLS}} \in \mathbb{R}^{1 \times d}$ and $X^{visual}_{patch} \in \mathbb{R}^{N \times d}$ denote the extracted features from the \texttt{[CLS]} token and image patches, respectively. For patch-based classification, we use only the features obtained from patches, $X^{visual}_{patch}$. CLIP's text encoder is leveraged to compute the textual embeddings, $X^{text} \in \mathbb{R}^{C \times d}$, for the target $C$ classes. Then, we use cosine similarity to determine similarity scores between the text embeddings and patch features. After the softmax operation, we interpolate the resulting logit to resize it to the original image dimensions. Finally, we perform a class-wise argmax operation to produce the segmentation map $S \in \mathbb{R}^{H \times W}$. Formally, 

\begin{equation}
  S = \argmax_{c \in C}(upsample(cos(X^{visual}_{patch},X^{text})))
  \label{eq:eq_5}
\end{equation}

\noindent where $cos$ represents cosine similarity and $upsample$ denotes bilinear interpolation. Note that we omit the softmax for notational simplicity in this expression.

\begin{figure}[t]
  \centering
  \includegraphics[width=0.9\columnwidth]{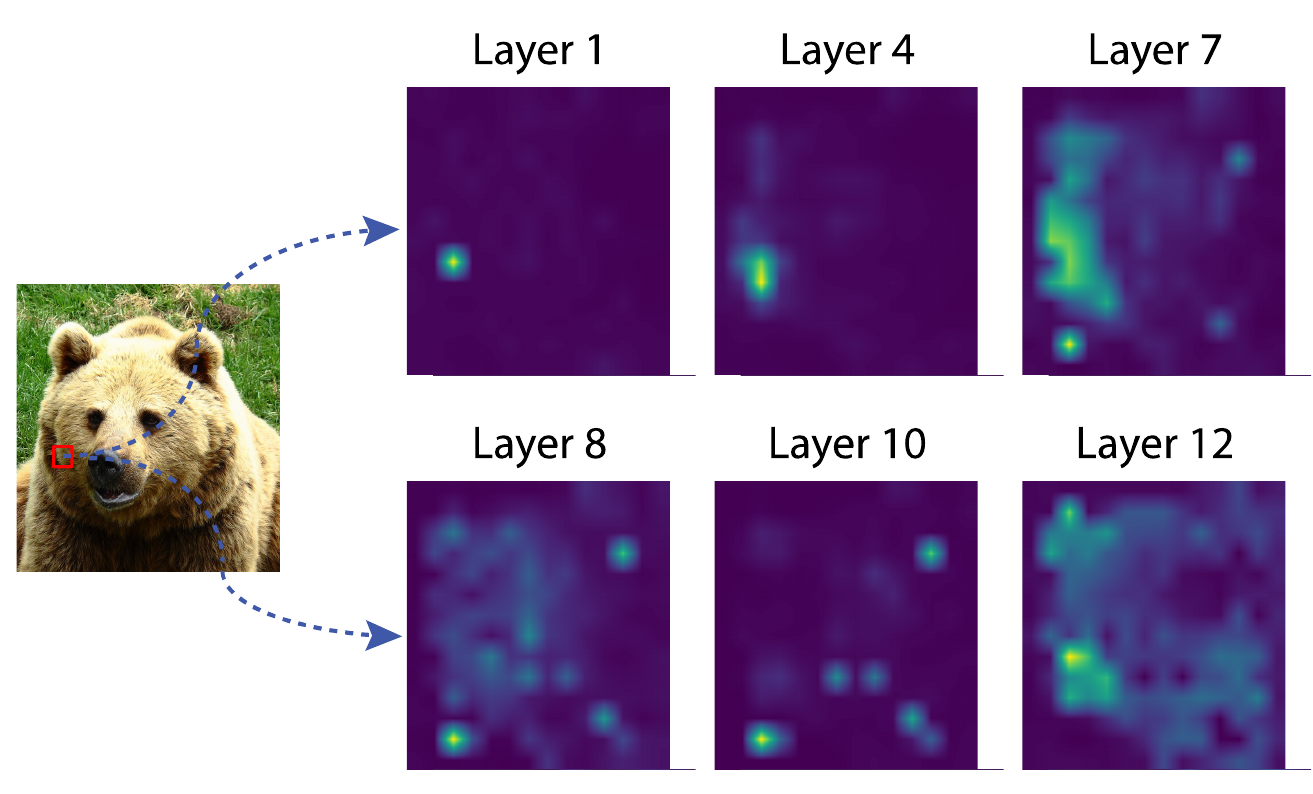}
  \caption{\textbf{Visualization of attention maps from various layers for a selected patch.} The red rectangle indicates the position of the randomly selected patch. Note that we use CLIP-ViT-B/16 as our visual backbone, with Layer 12 serving as the final layer.}
  \label{fig:fig_3}
\end{figure}

\subsection{Architectural Modifications}
\label{sec:arch_change}
\textbf{Self-Self Attention.} Recent studies \cite{wang2023sclip,hajimiri2024pay,bousselham2024grounding,lan2024clearclip,li2023clip} demonstrate that the vanilla implementation has failed to achieve accurate object segmentation. In order to strengthen CLIP's segmentation performance, these studies propose modifying the attention mechanism, resulting in substantial improvement over the vanilla approach. The original self-attention module in CLIP is designed for image-level predictions using $x_{\texttt{CLS}}$. Because patch tokens do not directly engage with the text features during training, the localization information extracted from attention maps is limited. It is argued that this lack of localization leads to CLIP's suboptimal performance in dense prediction tasks \cite{bousselham2024grounding,hajimiri2024pay}. Thus, we focus on the concept of self-self attention, \eg, key-key (k-k), and query-query (q-q) attentions. Self-self attentions ensure that visual tokens attend to themselves, resulting in higher values both on the diagonal of the attention map and in semantically correlated patches \cite{wang2023sclip}. Compared to the original attention mechanism, self-self attention produces a more spatially localized attention map. During the experiments, we observed that employing q-q and k-k attentions, similar to SCLIP, yields the best performance. Formally, we use the following attention map formula in the last self-attention block of the image encoder,

\begin{equation}
\begin{aligned}
  \texttt{\textbf{Attn}}(X) = softmax(\frac{XW_{Q}W_{Q}^{T}X^{T}} {\sqrt{d}}) \\ + \: softmax(\frac{XW_{K}W_{K}^{T}X^{T}} {\sqrt{d}}).
  \label{eq:eq_6}
\end{aligned} 
\end{equation}


\textbf{Removing Feed-Forward Block.} The original Transformer \cite{vaswani2017attention} architecture includes FFN layers that process the output of the self-attention block through a series of fully connected layers, enabling the model to capture the complex relationships in the data. ViT used in CLIP's image encoder also incorporates this feed-forward block. However, due to the inherent structure of pre-training, the parameters of the FFN are optimized for tasks at the image level. Furthermore, CLIPSurgery \cite{li2023clip} finds that the FFN in the last encoder block negatively impacts CLIP's segmentation performance. Therefore, we propose discarding the feed-forward block in the final layer. Our new residual attention block in this layer can be expressed as,

\begin{equation}
  X^{(L)} = X^{(L-1)} + \texttt{{\textbf{SA}}}(LN(X^{(L-1)}))
  \label{eq:eq_7}
\end{equation}

\noindent where $L$ represents the total number of layers, and \texttt{{\textbf{SA}}} denotes the self-attention module that utilizes the modified attention map formula presented in \cref{eq:eq_6}.

\textbf{Incorporating Middle Layer Attention Maps with the Final Layer.} Previous methods rely solely on the final layer's attention map while ignoring valuable feature representations found in intermediate layers. A recent study \cite{gandelsmaninterpreting} demonstrates that attention heads in different layers, particularly in the last few layers, are specialized for various image properties (\eg, one head specializes in shapes, while another focuses on colors). Neglecting encoded information in intermediate layers hinders CLIP from achieving its full potential. 

In light of these findings, we analyze the attention maps extracted from different layers by applying the modified attention formula described in \cref{eq:eq_6}. \cref{fig:fig_3} illustrates these attention maps across various layers for a selected patch. We can observe that attention maps from shallow layers tend to attend only to the given patch position. Nevertheless, as we progress from shallow to intermediate layers, attention maps begin to highlight semantically correlated regions related to the selected patch. Although attention maps from the layers just before the final layer may lose certain spatial details, the attention map of the final layer provides the most informative map for the given patch (\eg, the shape of the bear is clearly visible in the final layer). This observation explains why methods that rely solely on the final layer's attention map can achieve effective results. On the other hand, this observation also indicates that intermediate layers contain rich information about the image. To leverage this valuable knowledge, we compute two distinct attention maps: the first directly from the final layer and the second by averaging the attention maps from selected intermediate layers. Subsequently, we average these two maps to produce our refined attention map. Formally, 

\vspace{-1em}
\begin{equation}
  \texttt{\textbf{Attn}}(X) = (\frac{\texttt{\textbf{Attn}}_{L}(X) + mean(\texttt{\textbf{Attn}}_{l'}(X))} {2})
  \label{eq:eq_8}
\end{equation}

\noindent where $\texttt{\textbf{Attn}}_{L}(X)$ and $\texttt{\textbf{Attn}}_{l'}(X)$ represent the attention maps for input $X$ from the final layer $L$ and selected intermediate layers $l'$, respectively. $mean$ denotes the mean operation across layers, and $\texttt{\textbf{Attn}}(X)$ is our refined attention map. A detailed analysis of the selected intermediate layers is presented in \cref{sec:ablation}.

\subsection{LLM-based Auxiliary Text Generation}
\label{sec:sec_llm}
With the emergence of large foundational VLMs such as CLIP, models have begun to understand texts beyond the class names in a closed set, allowing for incorporating auxiliary texts alongside original class names. Furthermore, LaCLIP \cite{fan2023improving} leverages the In-Context Learning (ICL) \cite{brown2020language,min2022metaicl} capability of LLMs to rewrite image captions. It is then trained with these augmented texts, resulting in superior zero-shot classification performance compared to CLIP. Inspired by the success of LaCLIP, we introduce a systematic approach to generate auxiliary texts for open-vocabulary semantic segmentation. We argue that this systematic strategy is more suitable for open-vocabulary settings than manually selecting auxiliary texts since LLMs perform well with classes beyond the predefined set.

To fully leverage the text encoder of CLIP, we generate a synonym and a definition for each given class using LLaMa 3 \cite{dubey2024llama} due to its strong ICL capabilities and open-source nature. For each dataset, we select either synonyms or definitions as the auxiliary text type based on their segmentation performance. Subsequently, we utilize CLIP's text encoder for original class names and auxiliary texts to obtain text embeddings $X^{text}$ and $X^{text}_{aux}$, respectively. Lastly, we combine these two text embeddings through a weighted summation as follows,

\begin{equation}
  X^{text}_{refined} = \alpha X^{text}_{aux} + (1 - \alpha) X^{text}
  \label{eq:eq_11}
\end{equation}

\noindent where $X^{text}_{refined}$ represents the resulting text features, and $\alpha$ is the auxiliary text coefficient.

\begin{table*}
  \centering
  \begin{tabular}{l|c|ccccc}
    \toprule
    Method & Post-process & COCO-Stuff & COCO-Object & VOC & Context & Cityscapes  \\
    \midrule
    Baseline & - & 7.1 & 8.6 & 20.3 & 9.0 & 6.9 \\
    \midrule
    ReCo \cite{shin2022reco} & - & 14.8 & 15.7 & 25.1 & 19.9 & 21.6 \\
    GroupViT \cite{xu2022groupvit} & - & 15.3 & 27.5 & 52.3 & 18.7 & 18.5 \\
    TCL \cite{cha2023learning} & PAMR & 19.6 & 30.4 & 55.0 & 30.4 &  23.5 \\
    \midrule
    MaskCLIP \cite{zhou2022extract} & - & 14.6 & 20.6 & 38.8 & 23.2 & 24.9 \\
    CLIP-DIY \cite{wysoczanska2024clip} & - & - & 31.0 & 59.9 & - & -\\
    ClearCLIP \cite{lan2024clearclip} & - & 23.9 & 33.0 & 51.8 & 32.6 & 30.0 \\
    SCLIP \cite{wang2023sclip} & PAMR & 23.9 & 32.1 & 61.7 & 31.5 & 34.1 \\
    NACLIP \cite{hajimiri2024pay} & PAMR & \underline{25.7} & 36.2 & 64.1 & \underline{35.0} & \underline{38.3} \\
    TagCLIP$^{*}$ \cite{lin2024tagclip} & - & 18.7 & 33.5 & 64.8 & - & - \\
    CaR \cite{sun2024clip} & Dense-CRF & - & \underline{36.6} & \underline{67.6} & 30.5 & - \\
    ITACLIP (Ours) & PAMR & \textbf{27.0} & \textbf{37.7} & \textbf{67.9} & \textbf{37.5} & \textbf{40.2} \\
    \bottomrule
  \end{tabular}
  \caption{\textbf{Comparison of ITACLIP with state-of-the-art methods (mIoU, \%).} We indicate which post-processing method has been applied to each model, if applicable. $^{*}$ denotes our reimplementation of this model on the COCO-Stuff and COCO-Object datasets. Note that the original paper of TagCLIP \cite{lin2024tagclip} evaluates the model on 27 mid-level categories of COCO-Stuff rather than on all 171 classes. Hence, we re-evaluate TagCLIP on COCO-Stuff using all class names for a fair comparison. For each dataset, \textbf{bold} values highlight the best scores, while \underline{underlined} values signify the second-best scores. \vspace{-1em}}
  \label{tab:tab_1}
\end{table*}

\subsection{Image Engineering}
\label{image_eng}
Since CLIP is trained on complete captions, several studies \cite{guopen,zhou2022extract,cha2023learning,wysoczanska2024clip,lan2024clearclip} employ prompt templates—an ensemble of different prompts such as \texttt{"a photo of the \{class name\}."}—to input \textit{prompt-engineered} captions into CLIP's text encoder. This \textit{Prompt Engineering} strategy enables models to fully leverage the text encoder and improve performance, as demonstrated in \cite{guopen}. However, previous studies have not applied a similar approach to enhance the input representations for the image encoder, only feeding the original image. Therefore, we propose the \textit{Image Engineering} module, which incorporates data augmentation techniques to expand the input representation. Note that, from this point forward, we omit the \texttt{[CLS]} token to avoid redundancy, as the vanilla approach does not use the \texttt{[CLS]} token for segmentation (see \cref{vanilla}). Thus, $X^{visual}$ and its attention map can be represented as $X^{visual}_{patch} \in \mathbb{R}^{N \times d}$ and $\texttt{\textbf{Attn}}(X) \in \mathbb{R}^{N \times N}$, respectively.

We categorize image augmentations into two groups based on whether they change the spatial structure of the image. Augmentations in the first category maintain the spatial arrangement of the image unchanged, whereas second category augmentations alter the spatial order of the image, requiring a reversal of the augmentation to preserve spatial information. We apply 1) Gaussian blur and 2) grayscale augmentations for the first category, while 1) horizontal and 2) vertical flips are utilized for the second category. Thus, each input image generates four augmentations.

\begin{table}
  \centering
  \begin{tabular}{cc}
    \toprule
    Attention Combination & VOC \\
    \midrule
    q-k & 19.0 \\
    \midrule
    q-q & 58.9 \\
    k-k & 52.2 \\
    v-v & 57.7 \\
    q-q + k-k & \textbf{67.9} \\
    q-q + v-v & 64.9 \\
    q-q + k-k + v-v & 66.4 \\
    \bottomrule
  \end{tabular}
  \caption{\textbf{Self-self attention combinations.} We evaluate our method with different self-self attention combinations on Pascal VOC. v-v represents the value-value attention. \vspace{-1.5em}}
  \label{tab:tab_2}
\end{table}

We perform two separate calculations for each augmentation category. Employing our revised image encoder, we compute the first-category visual features $X_{1}^{visual}$ as,

{\small
\begin{equation}
  X_{1}^{visual} = \frac{1}{K+1} \sum_{i=0}^{K} \frac{X^{visual,i}_{1}} {\|X^{visual,i}_{1}\|_F}
  \label{eq:eq_9}
\end{equation}}

\noindent where $\|\; \|_F$ corresponds to the Frobenius norm across the $d$ dimension. $X_{1}^{visual,i}$ denotes the visual features from the $i^{th}$ first-category augmentation of a total of $K=2$ augmentations and $X_{1}^{visual,0}$ is the image itself.

For the second-category augmentations, we feed augmented images into the image encoder, excluding the original image itself this time. Since the spatial order of the image has been altered, we cannot directly incorporate second-category visual features with $X_{1}^{visual}$. Instead, we separately calculate the logits from first and second-category augmentations, namely $L_{1}$  and $L_{2}$. $L_{1} \in \mathbb{R}^{C \times H \times W}$ is obtained directly by multiplying the embeddings $X_{1}^{visual}$ and $(X^{text}_{refined})^{\mathsf{T}}$. We then apply a similar operation to all second-category augmented images, generating corresponding logits for each image. To restore the original spatial order, we reverse these augmentations (\eg, a horizontally flipped image is flipped horizontally again). After this reversal, we average the resulting logits to obtain second-category logits. $L_{2} \in \mathbb{R}^{C \times H \times W}$. Lastly, we combine $L_{1}$ with $L_{2}$ through a weighted summation. Formally,

\vspace{-0.5em}

\begin{equation}
  L = \lambda L_{1} + (1 - \lambda) L_{2}
  \label{eq:eq_10}
\end{equation}

\noindent where \mbox{$L \in \mathbb{R}^{C \times H \times W}$} represents the refined \textit{image-engineered} logits, and $\lambda$ is introduced as the image engineering coefficient. Subsequently, we generate the segmentation map $S$ using our refined logits, following the baseline method outlined in \cref{vanilla}.
\vspace{-0.5em}

\section{Experiments}
\subsection{Experimental Setup}
\textbf{Datasets.} For a fair comparison with previous studies, we evaluate our method on five common semantic segmentation benchmarks using their official validation sets: COCO-Stuff \cite{caesar2018coco}, COCO-Object \cite{lin2014microsoft}, Pascal Context \cite{mottaghi2014role}, Pascal VOC \cite{everingham2010pascal}, and Cityscapes \cite{cordts2016cityscapes}. Specifically, COCO-Stuff comprises 171 classes without an explicit background class. COCO-Object is derived from the COCO-Stuff dataset, combining all stuff classes in COCO-Stuff into a single background class and including 80 thing classes from the original dataset. Pascal Context consists of 59 categories and one background class. Pascal VOC contains 20 object classes in addition to one background class. The Cityscapes dataset focuses on urban street scenes, with the validation set containing 19 semantic categories and no separate background class. The validation splits of these datasets include 5000, 5000, 5104, 1449, and 500 images, respectively. We assess the segmentation performance of our method using the mean Intersection over Union (mIoU) metric.

\begin{table}
  \centering
  \begin{tabular}{>{\centering\arraybackslash}m{0.8cm}|ccccc}
    \toprule
    FFN & Stuff & Object & VOC & Context & City \\
    \midrule
    \ding{51} & 26.3 &  36.9 &  66.3 &  36.3 &  39.4 \\
    \ding{55} & \textbf{27.0} &  \textbf{37.7} &  \textbf{67.9} &  \textbf{37.5} & \textbf{40.2} \\
    \bottomrule
  \end{tabular}
  \caption{\textbf{Removing the feed-forward block.} ``Stuff'', ``Object'', and ``City'' refer to COCO-Stuff, COCO-Object, and Cityscapes. \vspace{-1.5em}}
  \label{tab:tab_3}
\end{table}

\textbf{Implementation Details.} We use the CLIP-ViT-B/16 \cite{radford2021learning} model with OpenAI's pre-trained weights. We define a set of potential background classes for the Pascal VOC and COCO-Object datasets, similar to previous works \cite{wang2023sclip, sun2024clip, lin2024tagclip}. This set is fed into CLIP's text encoder to generate background representations. Additional details about the background set are provided in the Appendix. We utilize ImageNet \cite{deng2009imagenet} prompt templates used in CLIP to construct \textit{prompt-engineered} texts. Following previous studies \cite{wang2023sclip,sun2024clip,hajimiri2024pay}, we apply the text expansion technique that uses additional captions for specific classes (\eg, for ``person'' class: person, person in shirt, person in dress). Furthermore, we perform slide inference with a 224 × 224 window and a stride of 28, after resizing the short side of images to 336 (560 for
Cityscapes), following the procedure described in \cite{wang2023sclip}. Finally, we employ Pixel-Adaptive Mask Refinement (PAMR) \cite{araslanov2020single} to reduce noise in our predictions.

\textbf{Baselines.} We adopt the vanilla approach described in \cref{vanilla} as our baseline method, preserving the original implementation details while omitting post-processing. Also, we compare ITACLIP with other training-free semantic segmentation models, including MaskCLIP \cite{zhou2022extract}, CLIP-DIY \cite{wysoczanska2024clip}, ClearCLIP \cite{lan2024clearclip}, SCLIP \cite{wang2023sclip}, NACLIP \cite{hajimiri2024pay}, CaR \cite{sun2024clip}, and TagCLIP \cite{lin2024tagclip}, along with weakly-supervised approaches, namely ReCo \cite{shin2022reco}, GroupViT \cite{xu2022groupvit} and TCL \cite{cha2023learning}. 

\subsection{Main Results}

\cref{tab:tab_1} presents the segmentation performance of ITACLIP compared to other approaches. ITACLIP outperforms current state-of-the-art (SoTA) methods on the COCO-Stuff, COCO-Object, Pascal Context, and Cityscapes datasets by a notable margin. In addition, it achieves state-of-the-art performance on Pascal VOC, though with a relatively narrow margin. Furthermore, we emphasize that our method is more robust across all five datasets, whereas most other models experience a performance drop on at least one dataset. For instance, the CLIP-ES-based \cite{lin2023clip} methods listed in this table—CaR \cite{sun2024clip} and TagCLIP \cite{lin2024tagclip}—fail to maintain their segmentation capability on datasets containing numerous ``stuff'' classes (\eg, see the results of CaR on Pascal Context and TagCLIP on COCO-Stuff). This performance degradation limits the applicability of these models in real-world scenarios.

In addition to training-free models, we compare our approach with weakly-supervised models, including GroupViT \cite{xu2022groupvit} and TCL \cite{cha2023learning}. We observe that ITACLIP even outperforms these weakly-supervised models without any supervision. We also report the baseline model's scores in \cref{tab:tab_1} to illustrate the significant performance gap and demonstrate the need for more advanced approaches. 

\begin{table}
  \centering
  \begin{tabular}{cc}
    \toprule
    Intermediate Layers ($l'$) & VOC \\
    \midrule
    \ding{55} & 65.0 \\
    \midrule
    \{7\} & 65.4 \\
    \{8\} & 65.5 \\
    \{7, 8\} & 65.5 \\
    \{7, 8, 10\} & \textbf{65.6} \\
    \{7, 8, 9, 10\} & 65.5 \\
    \bottomrule
  \end{tabular}
  \caption{\textbf{Impact of selected intermediate layers.} We assess ITACLIP with various intermediate layers on Pascal VOC. \ding{55} indicates that the model does not use intermediate layers for evaluation.}
  \label{tab:tab_4}
\end{table}


\begin{table}
  \centering
  \begin{tabular}{c|ccccc}
    \toprule
    PAMR & Stuff & Object & VOC & Context & City\\
    \midrule
    \ding{55} & 26.3 &  36.4 &  65.6 &  36.0 & 39.2 \\
    \ding{51} & \textbf{27.0} &  \textbf{37.7} &  \textbf{67.9} &  \textbf{37.5} & \textbf{40.2} \\
    \bottomrule
  \end{tabular}
  \caption{\textbf{Effect of post-processing operation.} Comparing the performance of ITACLIP with and without PAMR on all datasets. \vspace{-2.5em}}
  \label{tab:tab_6}
\end{table}

\subsection{Ablation Study}
\label{sec:ablation}
\textbf{Self-self attention combinations.} Self-self attentions yield a more spatially localized attention map, as detailed in \cref{sec:arch_change}. To identify the optimal self-self attention combination, we perform an ablation study as presented in \cref{tab:tab_2}. Our method performs best using q-q and k-k attentions with a sum operation. We also provide the score of the original attention formula (q-k attention) to underscore the necessity of self-self attention. 

\textbf{Removing the feed-forward block.} In \cref{tab:tab_3}, we investigate the effect of removing the feed-forward block. We observe that FFN in the last layer leads to a drop in segmentation performance across all datasets. The results support the rationale detailed in \cref{sec:arch_change} for removing the FFN.

\textbf{Study on selected intermediate layers ($l'$).} We analyze the impact of various layers to determine which intermediate layers yield the best performance; \cref{tab:tab_4} summarizes our findings. To clearly observe the effect of layers, we evaluate the model without PAMR. Based on the visualization in \cref{fig:fig_3} and a recent study \cite{gandelsmaninterpreting} showing that attention maps from the last few layers have a greater effect on the representation, we concentrate more on the middle-deep layers in our analysis. Empirically, employing the 7th, 8th, and 10th layers as intermediate layers results in superior performance. We also report the score using only the final layer's attention map to demonstrate the effectiveness of our multi-layer approach.


\begin{table}
  \centering
  \begin{tabular}{lcc|cc}
    \toprule
     Method & LTG & IE & Context & Stuff \\
    \midrule
     ITACLIP & \ding{55} & \ding{55} & 34.3 & 24.5 \\
     ITACLIP & \ding{51} & \ding{55} & 34.6 & 24.8 \\
     ITACLIP & \ding{51} & \ding{51} & \textbf{35.4} & \textbf{25.4} \\
    \bottomrule
  \end{tabular}
  \caption{\textbf{Effect of Image Engineering and LLM-based Text Generation modules.} LTG and IE represent the LLM-based Text Generation and Image Engineering modules, respectively. \vspace{-1em}}
  \label{tab:tab_5}
\end{table}

\textbf{Effect of Image Engineering and LLM-based Text Generation modules.} We conduct an ablation study to assess the contributions of the Image Engineering and LLM-based Auxiliary Text Generation modules, as illustrated in \cref{tab:tab_5}. To better understand the impact of these modules, we evaluate our method without applying PAMR. ITACLIP demonstrates superior performance when both modules are employed. This result validates the hypothesis that enriched input features, derived from the Image Engineering and LLM-based Auxiliary Text Generation modules, can enhance image segmentation performance.

\textbf{Effect of post-processing operation.} Before generating the final predictions, we apply PAMR to mitigate noise in our results. We examine the effect of PAMR on our method across all datasets, as shown in \cref{tab:tab_6}. Although PAMR improves segmentation performance across all datasets, ITACLIP still achieves competitive results and even surpasses the SoTA on some datasets without PAMR. These results demonstrate that our method is highly robust and performs well without relying heavily on post-processing.


\begin{table}
  \centering
  \begin{tabular}{>{\centering\arraybackslash}m{0.8cm}|ccccc}
    \toprule
    Stride & Stuff & Object & VOC & Context & City \\
    \midrule
     224 & 25.3 &  36.7 &  66.1 & 36.9 & 37.7 \\
    112 & 26.6 &  37.4 &  67.1 & 37.4 & 39.7 \\
    56 & 26.9 &  \textbf{37.7} &  \textbf{67.9} &  \textbf{37.5} & \textbf{40.2} \\
    28 & \textbf{27.0} &  \textbf{37.7} &  \textbf{67.9} & \textbf{37.5} & \textbf{40.2} \\
    \bottomrule
  \end{tabular}
  \caption{\textbf{Role of stride value.} We investigate the role of the stride value in our method across all five datasets.}
  \label{tab:tab_7}
\end{table}

\textbf{Role of stride value.} In \cref{tab:tab_7}, we investigate the impact of different stride values on our method. Our results indicate that lower stride values tend to improve segmentation performance. Considering that lower stride values increase computational cost, we also report ITACLIP's segmentation performance using higher stride values to enable faster inference. We observe that ITACLIP's performance with a stride of 56 nearly matches the results obtained using our default stride of 28. Even with a stride of 112, our method achieves state-of-the-art results on most datasets.

\begin{table}
  \centering
  \begin{tabular}{cc|c}
    \toprule
     $\lambda$ & $\alpha$ & Context \\
    \midrule
     0.75 & 0.15 & \textbf{36.0} \\
     0.6 & 0.15 & 35.9 \\
     0.5 & 0.15 & 35.8 \\
     0.75 & 0.2 & 35.9 \\
     0.6 & 0.2 & 35.9 \\
     0.6 & 0.1 & 35.9 \\
     0.5 & 0.1 & 35.8 \\
    \bottomrule
  \end{tabular}
  \caption{\textbf{Influence of Hyperparameters.} $\lambda$ and $\alpha$ denote the Image Engineering and Auxiliary Text Coefficients, respectively. The experiments are conducted without applying PAMR. \vspace{-1em}}
  \label{tab:tab_8}
\end{table}

\textbf{Influence of Hyperparameters.} \cref{tab:tab_8} illustrates the effect of Image Engineering ($\lambda$) and Auxiliary Text Coefficients ($\alpha$) on the Pascal Context dataset. We observe minimal performance variance within the range of values listed in the table, highlighting the robustness of our method.
\section{Conclusion}
This study introduces ITACLIP, which leverages CLIP’s image-level knowledge for dense prediction tasks without requiring pixel-level annotations or additional training. We propose architectural modifications to CLIP's visual encoder and introduce the Image Engineering and Auxiliary Text Generation modules to expand the input representation. As a result of these modifications and modules, our experiments show that ITACLIP achieves SoTA results across five segmentation datasets. Additionally, the Image Engineering module and LLM-based Text Generation strategy can be seamlessly integrated into a range of computer vision tasks. These modules will assist researchers in further improving prediction quality across various vision tasks.

{\small
\bibliographystyle{ieee_fullname}
\bibliography{egbib}
}

\clearpage
\newpage
\appendix
\FloatBarrier
\section{Appendix}
\subsection{Additional Implementation Details}
\label{app:app_1}
\begin{figure}[h]
    \centering
    \begin{subfigure}[b]{0.47\textwidth}
        \centering
        \includegraphics[width=\textwidth]{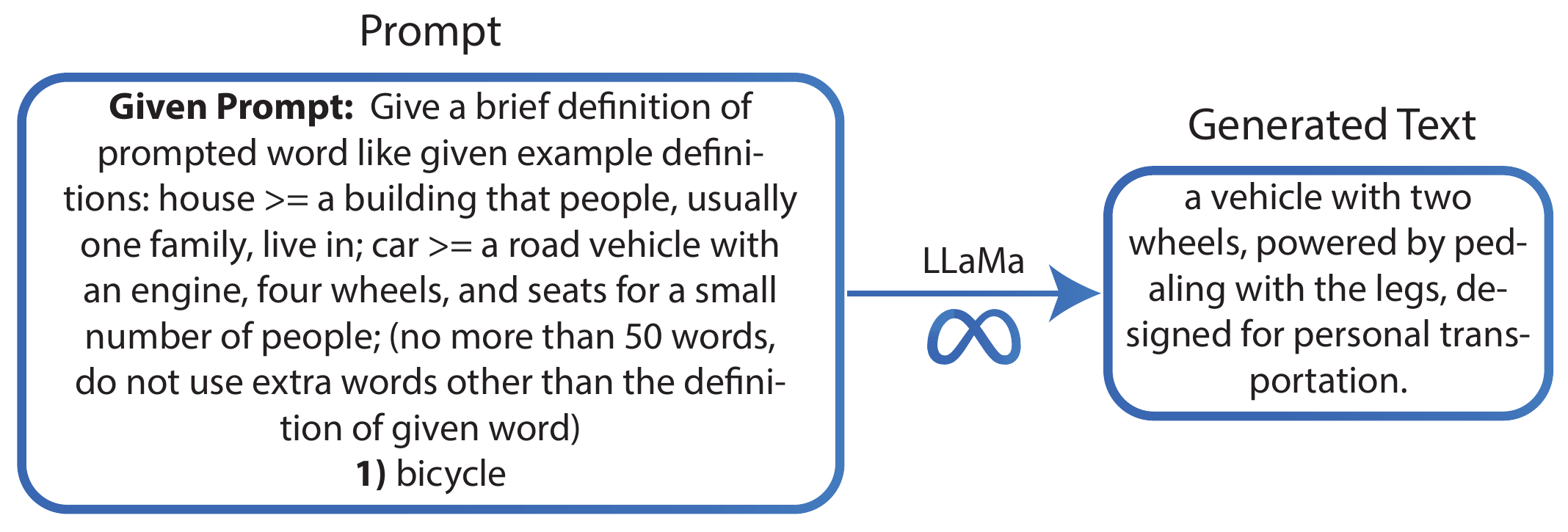}
        \caption{We employ the illustrated prompt to generate definitions.}
        \label{fig:part_4a}
    \end{subfigure}
    \hfill
    \begin{subfigure}[b]{0.47\textwidth}
        \centering
        \includegraphics[width=\textwidth]{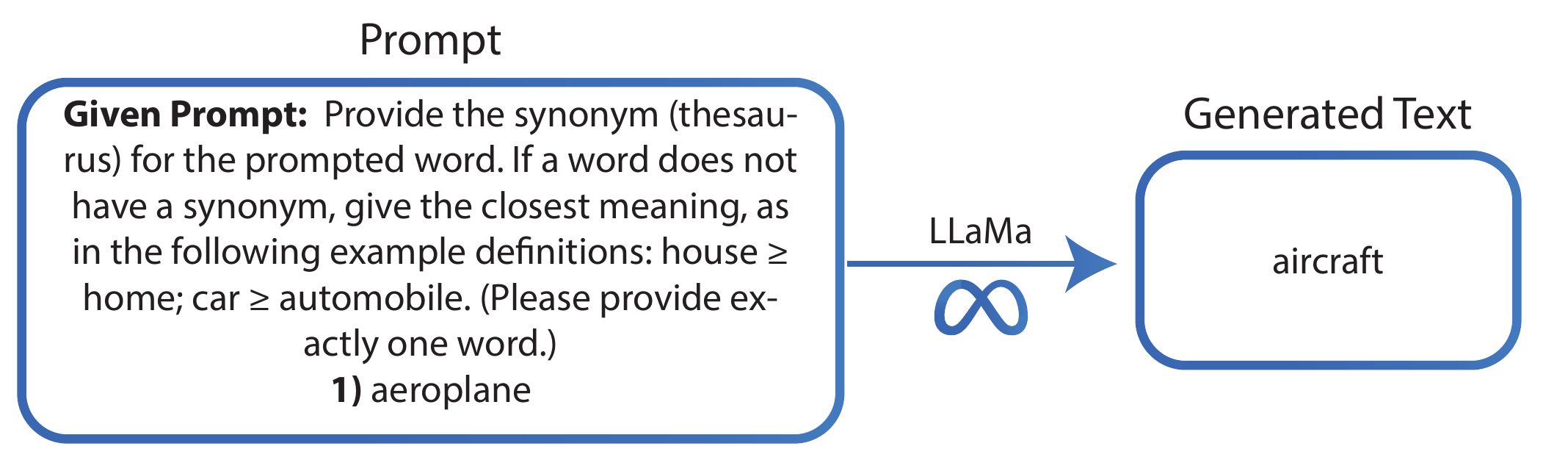}
        \caption{We employ the illustrated prompt to generate synonyms.}
        \label{fig:part_4b}
    \end{subfigure}
    \caption{\textbf{Procedure for generating auxiliary texts for a given class name.}}
    \label{fig:common_caption}
\end{figure}

\textbf{LLM-based Auxiliary Text Generation.} As detailed in \cref{sec:sec_llm}, we leverage LLama 3 \cite{dubey2024llama} to generate auxiliary texts for each class name. We use generated definitions as the auxiliary text type for the COCO-Stuff \cite{caesar2018coco}, Pascal VOC \cite{everingham2010pascal}, and Pascal Context \cite{mottaghi2014role} datasets and synonyms for the COCO-Object \cite{lin2014microsoft} and Cityscapes \cite{cordts2016cityscapes} datasets. \cref{fig:part_4a} and \cref{fig:part_4b} illustrate the procedure for generating definitions and synonyms, respectively. Example definitions and synonyms from the Cambridge Dictionary \cite{cambridge_dictionary} are utilized to guide LLaMa in producing more precise definitions.

\textbf{Image Engineering ($\lambda$) and Auxiliary Text Coefficients ($\alpha$).} The Image Engineering ($\lambda$) and Auxiliary Text Coefficients ($\alpha$) are employed in weighted summations within the Image Engineering and LLM-based Auxiliary Text Generation modules, respectively. \cref{tab_app_1} reports these hyperparameter values used across all evaluated datasets. As demonstrated in \cref{tab:tab_8}, the effects of $\lambda$ and $\alpha$ are insignificant within the range specified in the main paper, and our default values exhibit minimal variance across datasets.

\textbf{Background set.} Following previous studies \cite{wang2023sclip, sun2024clip, lin2024tagclip}, we define a background set for the Pascal VOC and COCO-Object datasets to enable our method to distinguish foreground classes from the background. Specifically, the background set employed in COCO-Object is

\texttt{background = } [\textit{sky, wall, tree, wood, grass, road, sea, river, mountain, sands, desk, bed, building, cloud, lamp, door, window, wardrobe, ceiling, shelf, curtain, stair, floor, hill, rail, fence}].

A similar background set is used for the evaluation of Pascal VOC.

\begin{table}
  \centering
  \begin{tabular}{c|ccccc}
    \toprule
    Hyperparameter & Stuff & Object & VOC & Context & City \\
    \midrule
    $\lambda$  & 0.75 &  0.75 &  0.7 &  0.75 & 0.7 \\
    $\alpha$ & 0.2 &  0.1 &  0.05 &  0.15 & 0.05 \\
    \bottomrule
  \end{tabular}
  \caption{\textbf{Hyperparameter values used in our experiments.}}
  \label{tab_app_1}
\end{table}

\subsection{Additional Experiments}

\textbf{Impact of different visual backbones.} We perform an ablation study to assess the impact of various CLIP-ViT backbones, as shown in \cref{tab_app_2}. ITACLIP achieves peak performance using the ViT-B/16 backbone, consistent with prior works \cite{lan2024clearclip,hajimiri2024pay}. 

\textbf{Background set.} In \cref{tab_app_3}, we analyze the effect of defining the background set on the COCO-Object dataset. Since the COCO-Object dataset consists of 80 ``thing'' classes and one explicit background class, our method fails to distinguish foreground classes from the background when the word ``background'' is solely used to define all possible background classes. We observe a substantial performance boost when a separate background set is defined.

\begin{table}
  \centering
  \begin{tabular}{cc}
    \toprule
    Backbone & VOC \\
    \midrule
    ViT-L/14 & 53.3 \\
    ViT-B/32 & 56.7 \\
    ViT-B/16 & \textbf{67.9} \\
    \bottomrule
  \end{tabular}
  \caption{\textbf{Impact of different visual backbones.} We compare the performance of ITACLIP with different visual backbones. }
  \label{tab_app_2}
\end{table}

\begin{table}
  \centering
  \begin{tabular}{cc|c}
    \toprule
    Method & Background Set & Object \\
    \midrule
    ITACLIP  & \ding{55} &  34.5 \\
    ITACLIP & \ding{51} &  \textbf{37.7}  \\
    \bottomrule
  \end{tabular}
  \caption{\textbf{Effect of background set.} ITACLIP performs better when the background set is employed.}
  \label{tab_app_3}
\end{table}

\subsection{More Qualitative Results}
\begin{figure*}
    \centering
    \includegraphics[width=1\linewidth]{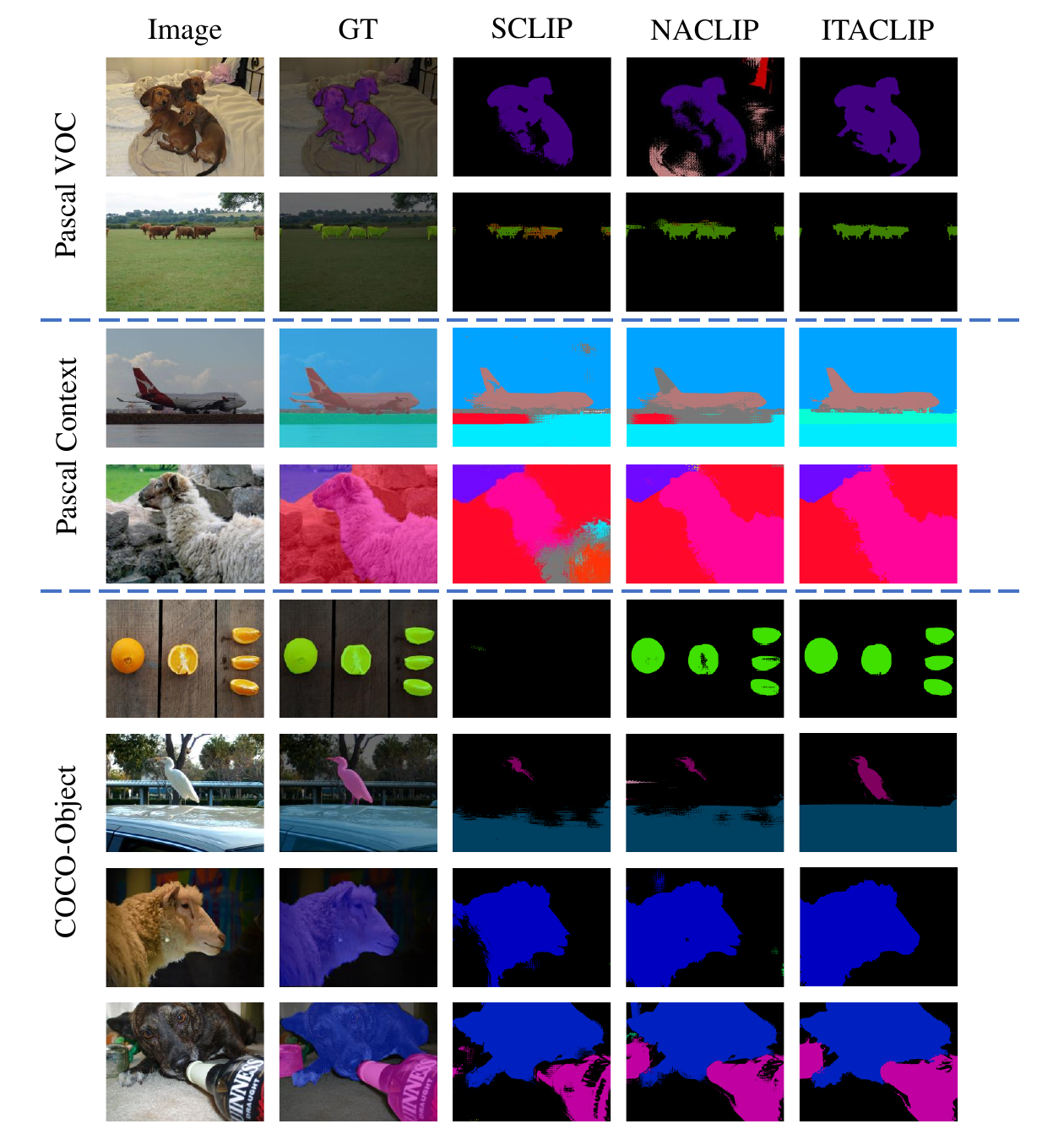}
    \caption{\textbf{Qualitative comparison of training-free semantic segmentation methods.} We compare ITACLIP with SCLIP \cite{wang2023sclip} and NACLIP \cite{hajimiri2024pay} using images from the Pascal VOC \cite{everingham2010pascal}, Pascal Context \cite{mottaghi2014role}, and COCO-Object \cite{lin2014microsoft} datasets. ITACLIP consistently outperforms the other approaches. GT denotes the ground truth of the image. }
    \label{fig:fig_5}
\end{figure*}

\cref{fig:fig_5} presents additional visualizations of ITACLIP on the COCO-Object, Pascal Context, and Pascal VOC datasets, comparing our method with SCLIP \cite{wang2023sclip} and NACLIP \cite{hajimiri2024pay}. As shown in \cref{fig:fig_5}, ITACLIP produces clearer segmentation masks compared to SCLIP and NACLIP, whose predictions are generally noisier. Furthermore, SCLIP and NACLIP occasionally fail to recognize objects accurately and predict classes not present in the image.

\end{document}